\documentclass[runningheads]{llncs}

\usepackage{eccv}

\usepackage{eccvabbrv}

\usepackage{graphicx}
\usepackage{multirow}
\usepackage{booktabs}
\usepackage{makecell}
\usepackage{amsmath}
\usepackage{bm}

\usepackage[accsupp]{axessibility}  

\usepackage{hyperref}

\usepackage{orcidlink}

\begin{document}

\title{Enhancing Lightweight Vision Language Models through Group Competitive Learning for Socially Compliant Navigation} 

\titlerunning{Abbreviated paper title}

\author{Xinyu Zhang\inst{1} \and
Atsushi Konno\inst{1} \and
Toshihiko Yamasaki\inst{2} \and
Ling Xiao\inst{1}$^{\dagger}$}%


\institute{
Hokkaido University, Sapporo, Japan\\
\and
The University of Tokyo, Tokyo, Japan\\
$^{\dagger}$Corresponding author: \texttt{ling@ist.hokudai.ac.jp}
}


\maketitle

\begin{abstract}
Social robot navigation requires a sophisticated integration of scene semantics and human social norms. Scaling up Vision Language Models (VLMs) generally improves reasoning and decision-making capabilities for socially compliant navigation. However, increased model size incurs substantial computational overhead, limiting suitability for real-time robotic deployment. Conversely, lightweight VLMs enable efficient inference but often exhibit weaker reasoning and decision-making performance in socially complex environments. Achieving both strong reasoning ability and efficiency remains an open challenge. To bridge this gap, we propose Group Competitive Learning (GCL), a strategy designed to amplify the capabilities of lightweight VLMs. Our strategy introduces the Group Competitive Objective (GCO) to harmonize global semantics with distributional regularization, alongside Asymmetric Group Optimization (AGO) to explore the upper limits of model performance.
Empirical evaluations on social navigation benchmarks demonstrate that GCL significantly elevates VLM performance. Specifically, GCL enables the Qwen2.5-VL-3B learner model and guide Qwen3-VL-4B to achieve an F1 score of 0.968 and 0.914, representing 40\% and 12\% improvement over vanilla supervised fine-tuning (SFT).  Notably, under vanilla SFT, the 3B model initially trails the 8B model (F1: 0.692 vs. 0.755). However, through the GCL, the 3B model outperforms (28\%) the 8B baseline model. These results suggest that GCL provides an effective solution for achieving both high accuracy and computational efficiency in real-world deployment.
\keywords{Social robot navigation \and small vision language models \and competitive learning \and human robot interaction}
\end{abstract}

\section{Introduction}
\label{sec:intro}
Social robot navigation is a fundamental challenge in embodied AI, requiring agents to not only perceive complex scene semantics but also strictly adhere to human social norms to generate compliant action commands~\cite{mavrogiannis2023core,raj2024rethinking}. The emergence of Vision Language Models (VLMs) offers a promising paradigm for enabling robots to reason about visual observations through semantic and linguistic grounding. By aligning perception with high-level language representations, VLMs facilitate more informed and socially aware navigation behaviors. 
Generally, increasing the scale of VLMs improves their reasoning and decision-making capabilities, which is particularly beneficial for socially compliant navigation tasks. However, larger models incur substantial computational cost, limiting their deployment in real-time robotic systems~\cite{huang2025multimodal,zhang2024interactive,batool2025impedancegpt}. Conversely, lightweight VLMs enable efficient inference but often struggle to capture the complex social cues required for robust navigation in crowded or dynamic environments~\cite{song2024vlm,payandeh2025social,kawabata2025socialnav}. Achieving a balance between strong reasoning capability and computational efficiency therefore remains a key challenge.

To bridge the gap between VLM performance and efficiency, Knowledge Distillation (KD) has been the dominant paradigm~\cite{elnoor2025vi,zhao2024experience,qian2025compressing}. Certain KD approaches focus on compressing the visual encoder or aligning cross-modal embeddings. Because, operating primarily at the visual feature level, these methods inadvertently neglect high-level reasoning consistency and specific decision-making processes, which are critical capabilities for social robot navigation tasks. Conversely, other techniques target the language model of VLMs by conducting distillation over the token-level output distributions~\cite{harish2024reinforcement, li2025robo}. Although these strategies improve the efficiency of knowledge transfer between models, them lack explicit coordination between global semantic reasoning and token-level behaviors. Consequently, directly leveraging these KD strategies to train lightweight VLMs remains fundamentally infeasible in practical social robotics scenarios constrained by severely limited onboard memory and restricted task-specific dataset sizes.

This paper aims to address these challenges. Specifically, we propose Group Competitive Learning (GCL), a novel strategy that enhances VLMs through group based competitive optimization, enabling effective knowledge transfer among models with similar or heterogeneous architectures without being constrained by model capacity. Our approach introduces three key contributions:


\begin{itemize}
    \item We propose Group Competitive Learning (GCL), a training strategy that can be seamlessly integrated into various VLMs. GCL improves socially compliant reasoning and action generation across both homogeneous and heterogeneous architectures, regardless of model capacity.
    
    \item We introduce a Group Competitive Objective (GCO) to align VLMs at both semantic and token distribution levels through semantic consistency and distributional regularization. We further propose Asymmetric Group Optimization (AGO) to reveal the optimization dynamics and performance boundaries of GCL while avoiding exhaustive hyperparameter search.
    
    \item Extensive experiments on two social navigation benchmarks show clear improvements over supervised fine-tuning (SFT). Under GCL, Qwen2.5-VL-3B improves by \textbf{40\%} (F1: 0.968) and Qwen3-VL-4B by \textbf{12\%}. Notably, although the base 3B model initially lags behind the 8B counterpart (0.692 vs. 0.755), after GCL it surpasses the 8B baseline with a 28\% improvement.
\end{itemize}

\section{Related Works}

\subsection{VLMs in Social Robot Navigation}
In human environments, social robot navigation~\cite{fang2026obstacles} refers to the robot’s ability to perceive its surroundings, interpret high-level semantic cues (e.g., human activities, spatial arrangements, social norms), and adjust its behavior in accordance with unwritten social conventions, such as maintaining personal space, yielding right-of-way, or avoiding collision trajectories through crowds~\cite{bae2025social,howard2024socialcounterfactuals}. Unlike traditional reactive or rule-based navigation, social robot navigation leverages multimodal perception and semantic reasoning to generate socially compliant motion actions that are both safe and natural from a human perspective~\cite{huang2025multimodal,song2024vlm,huang2025visbias}. This paradigm is particularly critical for service and assistive robots operating in dynamic, unstructured settings like hospitals, shopping malls, or homes.

While large VLMs offer strong perception and decision-making capabilities in human robot interaction tasks~\cite{narasimhan2025olivia,mukherjee2025toward,zhao2025vlm}, these models incur substantial computational overhead, limiting their deployment on resource constrained robotic platforms~\cite{payandeh2025social,kawabata2025socialnav,wang2025maction}. Consequently, there remains a critical challenge in exploring efficient, lightweight VLMs that retain sufficient reasoning capacity for social robot navigation tasks without sacrificing deployability.

\subsection{Knowledge Distillation in VLM}
The deployment of VLMs is often bottlenecked by their immense computational footprints. To address this, Knowledge Distillation (KD) has become a cornerstone for transferring capabilities to compact students~\cite{xiao2025multi,wu2022tinyvit,han2024amd}. Knowledge distillation encompasses two principal paradigms: online distillation and offline distillation. Certain approaches compress visual encoders or align cross-modal embeddings in VLMs, evaluated on image centric tasks (e.g., captioning, VQA)~\cite{jang2025vl2lite,li2024promptkd,naeem2024silc,de2025towards}. These methods operate predominantly at the visual feature level and neglect high-level reasoning consistency and embodied decision-making, critical capabilities for socially compliant robot navigation.

Several methods focus on matching in language models of VLMs~\cite{agarwal2024policy}, Shi et al~\cite{shi2025enhancing} augment the student model's reasoning capabilities by distilling the teacher model's chain-of-thought (CoT) outputs into experiential knowledge. Nevertheless, it remains inherently contingent upon both the teacher model's quality and the precision of token-level alignment. Fan et al~\cite{fan2025fedmkt} jointly optimize multiple models via prediction consistency using symmetric divergence losses and uniform schedules, showing efficacy in classification and language modeling~\cite{agarwal2024policy}. However, such strategies lack explicit coordination of global semantic reasoning and fine-grained token-level behaviors. When applied directly to VLMs with disparate capabilities and capacities, they often induce training instability or collapse, underscoring the need for capacity alignment mechanisms.

\section{Methods}

\subsection{Overview of Group Competitive Learning Strategy}
\label{sec:sub31}
\begin{figure*}[t]
    \centering
    \includegraphics[width=\linewidth]{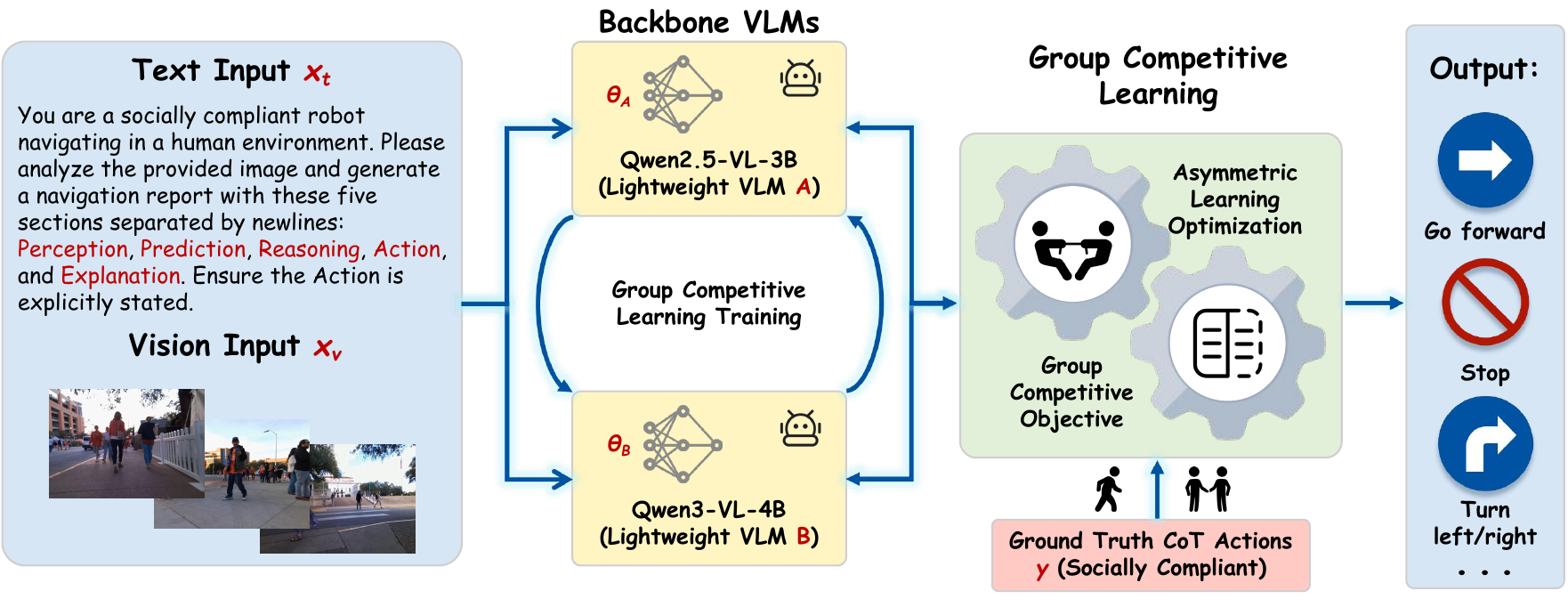}
    \caption{\textbf{The Overview of GCL.} The left of the figure illustrates the GCL strategy's inputs examples: texts and images across social navigation dataset. The middle shows the GCL strategy's components: Group Competitive Objective, to align VLMs at both semantic and distributional token levels, and Asymmetric Group Optimization, to explore the optimization dynamics and performance boundaries of GCL. The right shows the socially compliant instruction of GCL outputs.}
    \label{fig:method}
\end{figure*}

Existing VLM knowledge distillation approaches prioritize visual level alignment for image centric tasks. Directly adopting existing strategy tends to be oblivious to the high-level reasoning and embodied decision-making essential for socially compliant navigation. To address this, we propose the Group Competitive Learning (GCL), as shown in Fig.~\ref{fig:method}, aiming to construct a dynamic, dual stream synergistic strategy where two VLMs with distinct architectures and parameter scales can mutually enhance each other through benign competition.

Given a multimodal dataset $\mathcal{D} = \{(x_v, x_t, y)\}_{i=1}^N$ ($x_v$: visual input, $x_t$: text input, $y$: the target), GCL defines a competitive group $\mathcal{G} = \{A(\theta_A), B(\theta_B)\}$ ($\theta_A$, $\theta_B$: parameters of competitors models $A$, $B$). Coordinated interactions across feature-level semantic and token-level distributions, GCL maximizes the posterior  $P(y|x_v, x_t)$ while minimizing model semantic and generative discrepancies, thereby unlocking the group's collective potential. Specifically, GCL consists of two components: Group Competitive Objective (GCO) loss and Asymmetric Group Optimization (AGO). GCO promotes semantic and distributional token level alignment. AGO helps unleash the whole potential of GCL by matching the capability and capacity gap between the competitive group.

\subsection{Group Competitive Objective}
\label{sec:sub32}
To achieve comprehensive competition ranging from global semantic consistency to local generative precision, we design a loss function of Group Competitive Objective (GCO), $\mathcal{L}_{GCO}$, comprising three components:
\begin{equation}
    \mathcal{L}_{GCO} = \lambda_{sup}\mathcal{L}_{sup} + \lambda_{GSL}\mathcal{L}_{GSL} + \lambda_{DRL}\mathcal{L}_{DRL},
\end{equation}
where $\lambda_{sup} = 1.0, \lambda_{GSL} = 0.5, \text{and } \lambda_{DRL} = 0.4$ are the weights of each loss term.

\noindent\textbf{Supervised Loss.}
To preserve instruction following capabilities and maintain alignment with the task distribution throughout competitive training, we retain the autoregressive language modeling loss.  For any model $k \in \{A, B\}$ in the group, the loss is defined as:
\begin{equation}
    \mathcal{L}_{sup}^{(k)} = - \frac{1}{N} \sum\nolimits_{n=1}^{N} \log P(y_n | y_{<n}, x_v, x_t; \theta_k),
\end{equation}
where $n$ denotes the $n$-th step, $N$ represents the total number of steps. The total supervised loss is $\mathcal{L}_{sup} = \mathcal{L}_{sup}^{(A)} + \mathcal{L}_{sup}^{(B)}$.

\noindent\textbf{Global Semantic Loss (GSL).}
Existing online distillation approaches often neglect misalignment between different models when encoding high-level semantics feature. To address this, we propose the Global Semantic Loss (GSL). Since VLMs output variable length hidden state sequences $H \in \mathbb{R}^{L \times d}$, direct alignment is infeasible. GSL introduces a learnable attention pooling module to aggregate $H_A, H_B$ into a compact global semantic vector $z_A, z_B$.
Subsequently, we project the vectors into a common feature space with $L2$ normalization to obtain $\bar{z}_A, \bar{z}_B$. Finally, we employ an InfoNCE-style contrastive loss~\cite{yu2022adversarial} 
to maximize the semantic consistency of the $i$-th inputs across different models:
\begin{equation}
    \mathcal{L}_{GSL} = - \frac{1}{B_{size}} \sum_{i=1}^{B_{size}} \log \frac{\exp(\text{sim}(\bar{z}_{A,i}, \bar{z}_{B,i}) / \tau)}{\sum_{j=1}^{B_{size}} \exp(\text{sim}(\bar{z}_{A,i}, \bar{z}_{B,j}) / \tau)},
\end{equation}

Unlike pretraining, which demands extensive negative samples for representation learning from scratch, GSL operates during fine tuning and remains effective under memory constrained regimes with small global batch sizes (16). In GCL, GSL acts as a semantic direction guide, aligning the optimization direction of models with mature representations in feature space to enable high dimensional semantic competition and consensus.

\noindent\textbf{Distributional Regularization Loss (DRL).}
To address the sparsity of GSL’s feature-level alignment signals, we propose the Distributional Regularization Loss (DRL), a dense and batch independent regularization term operating directly on full vocabulary probability distributions over $\mathcal{V}$. We employ the Jensen-Shannon divergence to constrain the distributions of both models:
\begin{equation}
    \mathcal{L}_{DRL} = \mathcal{D}_{KL}(P_A || \mathcal{M})/2 + \mathcal{D}_{KL}(P_B || \mathcal{M})/2,
\end{equation}
where $P_A, P_B \in \mathbb{R}^{|\mathcal{V}|}$ denote the full softmax probability of the models, $\mathcal{D}_{KL}$ denote the Kullback-Leibler (KL) divergence and $\mathcal{M} = 1/2(P_A + P_B)$ represents the average distribution. 

As a soft constraint, DRL leverages token-level competition to transfer implicit knowledge through non-target token distributions, mitigating catastrophic degradation during fine-tuning and enhancing generalization to unseen distributions.

\subsection{Asymmetric Group Optimization}
\label{sec:sub33}
To resolve optimization dynamics mismatch and efficiently characterize GCL performance boundaries, we introduce Asymmetric Group Optimization (AGO), which decouples the joint parameter space into independent subspaces ($\Omega_A$, $\Omega_B$) for $\theta_A$ and $\theta_B$ of models A and B, assigning distinct learning rates $\eta$ and temperatures $\tau$ per model at step $t$:
\begin{equation}
\theta_{k}^{t+1} \leftarrow \text{Optimizer}(\nabla \mathcal{L}_{GCO}, \eta_k(t), \tau_k), \quad k \in \{A, B\}
\end{equation}
where $\theta{k}^{t+1}$ represents the updated parameters of model $k$, $\text{Optimizer}(\cdot)$ denotes the parameter update function, $\nabla \mathcal{L}_{GCO}$ is the gradient of the GCO loss, while $\eta_k(t) \neq \eta_{rival}(t)$ and $\tau_k \neq \tau_{rival}$ denote the model specific hyperparameters. Specifically, AGO introduces a dual assignment rule:

\noindent\textbf{Performance based Role Assignment (Learning Rate).} $\eta$ assignment is determined by SFT performance: the better performance model serves as the \textit{Guide}, assigned a lower learning rate ($\eta_{low}$) to preserve pretrained normative knowledge stability; the worse performance model functions as the \textit{Learner}, receiving a higher learning rate ($\eta_{high}$) to enhance plasticity for rapid alignment to ground truth (GT) and the Guide's semantic space.

\noindent\textbf{Capacity based Entropy Control (Temperature).} Temperature assignment is governed exclusively by parameter size, independent of model role. Larger models adopt $\tau_{low}=2.0$ to preserve discriminative feature boundaries and prevent entropy collapse; smaller models adopt $\tau_{high}=3.0$ to smooth output distributions and enhance plasticity for capturing ``dark knowledge''.

To rigorously demonstrate how this capacity based temperature scaling governs the learning dynamics, we analyze the gradient of our customized $\mathcal{L}_{DRL}$. Let $P_k,(k \in {A, B})$ denote the softened output probability distribution of model $k$. The normalized loss after adding the asymmetric temperature coefficient is defined as:
\begin{equation}
\mathcal{L}_{DRL} = \tau_A^2\mathcal{D}_{KL}(P_A || \mathcal{M})/2  + \tau_B^2\mathcal{D}_{KL}(P_B || \mathcal{M})/2
\end{equation}
where $\mathcal{M} = (P_A + P_B)/2$ denotes the mean distribution. By applying the chain rule with the temperature scaled softmax Jacobian, the gradient with respect to $z_{k,i}$ (the previous softmax logit of the $i$-th class for model $k$) reveals a profound asymmetry:
\begin{equation}
\partial \mathcal{L}_{DRL}/\partial z_{k,i} \propto \underbrace{\tau_k}_{\textbf{Basic Alignment}} - \underbrace{\left( \tau_{rival}^2 - \tau_k^2 \right)}_{\textbf{Asymmetric Shift Force}}
\end{equation}
\textbf{details provided in the Supplementary Material} where $\tau_{rival} \in {\tau_A, \tau_B} \setminus {\tau_k}$ represents the temperature of the other model in the competitive group. This mathematical formulation explicitly proves why temperature must be coupled with model size: The $\tau$ term exponentially amplifies the gradient stiffness of larger models ($\tau=2.0$), allowing them to act as rigid semantic anchors regardless of whether they are learning or guiding. Meanwhile, the discrepancy term $(\tau_{rival}^2 - \tau_k^2)$ injects a an attractive force for the larger model($\tau=2.0$) and dynamic repulsive force for the smaller model ($\tau=3.0$).

Through this AGO, the GCL strategy mathematically guarantees optimal synergistic gain, accommodating complex competitive groups (e.g., a smaller guide grouped with a larger learner) within a unified joint training process.

\section{Experiment}

\subsection{Experimental Setup}
\label{sec:sub41}   
\noindent\textbf{Datasets and Evaluation Metrics.} We conduct experiments using SNEI \cite{payandeh2025social} (265/60 train/test social navigation scenarios, 1,625 image-text pairs) and MUSON \cite{liu2025muson} (640/160 scenarios, 4,000 pairs) to evaluate performance and generalization. To provide a holistic assessment of the generated response quality, we employ a diverse set of semantic metrics:
we report Action-F1 score. Specifically, given an output action \(y\) and a ground truth \(g\), a pretrained BERT encoder~\cite{devlin2019bert} produces token embeddings \(\{y_e^1,\ldots,y_e^n\}\) and \(\{g_e^1,\ldots,g_e^m\}\), with $m$ and $n$ as the number of tokens in the \(y\) and \(g\), respectively. 
We calculate the cosine similarity $\mathrm{cosine}(y_e^j, g_e^k)$ between token pairs using the results of embedding. 
Based on the cosine similarity, recall \(R\) and precision \(P\) are computed as
\begin{align}
R = \frac{1}{m} \sum\nolimits_{k=1}^{m} \max_{1 \leq j \leq n} \mathrm{cosine}(y_e^j, g_e^k), \quad
P = \frac{1}{n} \sum\nolimits_{j=1}^{n} \max_{1 \leq k \leq m} \mathrm{cosine}(y_e^j, g_e^k),
\end{align} 
The Action-F1 is then obtained as the harmonic mean of precision and recall:
\begin{equation}
\text{Action-F1} = 2PR/(P + R),
\end{equation}
Action-F1 provides an overall balance.
To explicitly evaluate the high-level semantic alignment, a core objective of our GSL, we utilize Perception Cosine Similarity (Perception-cos) and Reasoning Cosine Similarity (Reasoning-cos). These two metric compute the cosine similarity between the sentence embeddings of the generated text and the reference using a pretrained Sentence-BERT model, offering a robust measure of semantic cosine similarity in embedding space.

\noindent\textbf{Model Baselines.} We define the competitive group $\mathcal{G} = \{A, B\}$, consisting of a Learner ($A$) and a Guide ($B$) to represent models of relatively weaker and stronger capabilities, respectively. GCL is benchmarked against vanilla supervised fine-tuning (SFT), where baselines are established by independently fine-tuning each model under identical data configurations to ensure a controlled comparison.

\noindent\textbf{Implementation Details.} 
All experiments are conducted on an 8$\times$ NVIDIA RTX8000 GPU cluster using DeepSpeed ZeRO-3 for memory efficiency. The optimizer is AdamW with $\lambda_{sup}=1.0$, $\lambda_{GSL}=0.5$, and $\lambda_{DRL}=0.4$. 
To evaluate both robustness and performance limits, we report the following configurations. 
For the SFT baseline, a fixed learning rate $\eta=1\times10^{-5}$ is used, representing independent model optimization. 
For ablation, symmetric DRL temperatures $\tau_A=\tau_B=2.0$ are adopted to create a challenging setting and verify the effectiveness of the loss functions under non-optimal optimization conditions. 
Under GCL training, two models are used: a learner and a guide. For fair comparison, the learner uses $\eta=1\times10^{-5}$ and the guide $\eta=5\times10^{-6}$.
We do not conduct physical robot experiments because our work focuses on the high-level planning layer of the navigation stack. In typical navigation systems, high-level planning, local trajectory generation, and low-level control are separated modules. 
Our VLM operates at the planning stage, generating socially compliant navigation intentions from multimodal perception. 
These decisions are executed by conventional local planners (e.g., DWA or TEB) and controllers producing velocity commands $(v,\omega)$. 
Thus, this study mainly improves reasoning and planning capabilities, a prerequisite before integration with physical robot platforms.

\begin{table*}[t]
\centering
\caption{\textbf{Performance of GCL across 12 competitive group.} $\eta_{guide}$ and $\eta_{learner}$ are allocated to SFT baseline superiority. The performance hierarchy from lowest to highest is: Qwen2.5-VL-3B (0.692) $<$ Qwen3-VL-8B (0.755) $<$ Qwen2.5-VL-7B (0.777) $<$ Qwen3-VL-2B (0.797) $<$ Qwen3-VL-4B (0.816). Meanwhile, temperature parameters ($\tau_L$, $\tau_G$) are assigned solely based on parameter size (larger capacity = 2.0, smaller capacity = 3.0). $F1$ denotes Action-F1. Bold text indicates optimal performance of the model.(\textbf{For MUSON results, details provided in the Supplementary Material})}
\label{tab:main}
\resizebox{\textwidth}{!}{
\begin{tabular}{l|c|c|c}
\toprule
\multicolumn{4}{l}{
\makecell[l]{\textbf{Regime A: Reshaping Mode} \\
\textbf{Capacity\textsubscript{L} < Capacity\textsubscript{G},\quad Small $\rightarrow$ Large,\quad $\bm{\eta_{learner} = 1 \times10^{-5}}$ $\bm{\eta_{guide} = 5 \times10^{-6}}$ }}} \\
\midrule
\textbf{Model Group (Learner / Guide)} & $\mathbf{\tau_L}$ / $\mathbf{\tau_G}$ & \textbf{Baseline ($F1_L$ / $F1_G$)} & \textbf{GCL ($F1_L$ / $F1_G$)} \\
\midrule
Qwen3-VL-2B / Qwen3-VL-4B    & \textbf{3.0 / 2.0} & 0.797 / 0.816 & \textbf{0.954} {\color{teal}(+0.157)} / 0.910 {\color{teal}(+0.094)} \\
Qwen2.5-VL-3B / Qwen3-VL-4B  & \textbf{3.0 / 2.0} & 0.692 / 0.816 & \textbf{0.968} {\color{teal}(+0.276)} / 0.914 {\color{teal}(+0.098)} \\
Qwen2.5-VL-3B / Qwen2.5-VL-7B& \textbf{3.0 / 2.0} & 0.692 / 0.777 & 0.869 {\color{teal}(+0.177)} / 0.915 {\color{teal}(+0.138)} \\
Qwen2.5-VL-3B / Qwen2.5-VL-8B& \textbf{3.0 / 2.0} & 0.692 / 0.755 & 0.964 {\color{teal}(+0.272)} / 0.811 {\color{teal}(+0.056)} \\
Qwen3-VL-2B / Qwen3-VL-2B& \textbf{3.0 / 2.0} & 0.797 / 0.797 & 0.841 {\color{teal}(+0.044)} / 0.884 {\color{teal}(+0.087)} \\
Qwen2.5-VL-3B / Qwen2.5-VL-3B& \textbf{3.0 / 2.0} & 0.692 / 0.692 & 0.793 {\color{teal}(+0.101)} / 0.876 {\color{teal}(+0.184)} \\
\midrule
\multicolumn{4}{l}{
\makecell[l]{\textbf{Regime B: Extraction Mode} \\
\textbf{Capacity\textsubscript{L} > Capacity\textsubscript{G},\quad Large $\rightarrow$ Small,\quad $\bm{\eta_{learner} = 1 \times10^{-5}}$ $\bm{\eta_{guide} = 5 \times10^{-6}}$ }}} \\
\midrule
\textbf{Model Group (Learner / Guide)} & $\mathbf{\tau_L}$ / $\mathbf{\tau_G}$ & \textbf{Baseline ($F1_L$ / $F1_G$)} & \textbf{GCL ($F1_L$ / $F1_G$)} \\
\midrule
Qwen2.5-VL-3B / Qwen3-VL-2B  & \textbf{2.0 / 3.0} & 0.692 / 0.797 & 0.949 {\color{teal}(+0.257)} / 0.941 {\color{teal}(+0.144)} \\
Qwen2.5-VL-7B / Qwen3-VL-2B  & \textbf{2.0 / 3.0} & 0.777 / 0.797 & 0.907 {\color{teal}(+0.130)} / 0.862 {\color{teal}(+0.065)} \\
Qwen3-VL-8B / Qwen3-VL-2B    & \textbf{2.0 / 3.0} & 0.755 / 0.797 & \textbf{0.884} {\color{teal}(+0.129)} / 0.945 {\color{teal}(+0.148)} \\
Qwen2.5-VL-7B / Qwen3-VL-4B  & \textbf{2.0 / 3.0} & 0.777 / 0.816 & 0.879 {\color{teal}(+0.102)} / \textbf{0.988} {\color{teal}(+0.172)} \\
Qwen3-VL-8B / Qwen3-VL-4B    & \textbf{2.0 / 3.0} & 0.755 / 0.816 & 0.795 {\color{teal}(+0.040)} / 0.976 {\color{teal}(+0.160)} \\
Qwen3-VL-8B / Qwen2.5-VL-7B  & \textbf{2.0 / 3.0} & 0.755 / 0.777 & 0.799 {\color{teal}(+0.044)} / \textbf{0.955} {\color{teal}(+0.178)} \\
\bottomrule
\end{tabular}
}
\end{table*}

\subsection{Main Results}
\label{sec:sub42}
We evaluate 12 competitive configurations spanning model architectures (Qwen2.5-VL~\cite{bai2025qwen2}, Qwen3-VL~\cite{yang2025qwen3}) and parameter scales from 2B to 8B. Table~\ref{tab:main} reveals a consistent optimization pattern across GCL strategies. The dual decoupled mechanism, comprising performance based learning rates and capacity based temperatures, robustly establishes two distinct competitive learning regimes.

\noindent\textbf{Regime A: The Reshaping Mode.} In cross capacity competitive learning, optimal performance mandates assigning a $\tau_{high}$ to the smaller model and a $\tau_{low}$ to the larger rival. For instance, the Qwen2.5-VL-3B / Qwen3-VL-4B group, assigning $\tau_{learner}=3.0$ and $\tau_{guide}=2.0$ achieves a SOTA Action-F1 of 0.968. Elevated temperature enhances the structural plasticity of the smaller VLM, enabling precise reshaping by the discriminative guidance of the large capacity guide. Simultaneously, a higher $\eta_{learner}$ accelerates latent space adaptation, ensuring synchronization with the guide’s stable updates under a lower $\eta_{guide}$.

\noindent\textbf{Regime B: The Extraction Mode.} Conversely, when a large capacity model serves as the learner grouped with a smaller guide, entropy dynamics invert: the large model adopts a $\tau_{low}$ to preserve token-level discriminative precision, while the small model employs a $\tau_{high}$ to yield a smoothed output distribution that filters rigid biases and reveals implicit knowledge. The elevated $\eta_{learner}$ accelerates assimilation of guide refined knowledge by the large capacity learner. The learner supplies sharp directives as a reference benchmark, enabling the guide to perform comparative error correction through \textit{checking its answers against the learner's}, thereby optimizing distributional outputs under resource constraints. This mechanism elevates the Qwen3-VL-4B baseline from 0.816 to 0.988 under guidance from Qwen2.5-VL-7B.

While the SFT phase did not involve exhaustive hyperparameter optimization, potentially leading to a discrepancy between model capacity and performance, our results highlight that GCL remains consistent performance gains regardless of the SFT training quality.
\begin{table}[t!]
\centering
\caption{\textbf{Performance comparison of GCL components and vanilla SFT.} We report Action-F1 (F1), Perception-cos (Per-cos) and Reasoning-cos (Rea-cos) on two datasets SNEI and MUSON. These results demonstrate that GCL, enables compact models to internalize intricate structural knowledge, ultimately surpassing the generalization capabilities of larger models trained in isolation.}
\label{tab:ab_main}
\resizebox{\columnwidth}{!}{%
\begin{tabular}{l|cc|c|ccc|ccc}
\toprule
\multirow{2}{*}{} & 
\multicolumn{2}{c|}{\multirow{2}{*}{\textbf{Method}}} & 
\multirow{2}{*}{\textbf{Model}} & 
\multicolumn{3}{c|}{\textbf{SNEI}} & 
\multicolumn{3}{c}{\textbf{MUSON}} \\
& & & & 
\textbf{F1$\uparrow$} & \textbf{Per-cos$\uparrow$} & \textbf{Rea-cos$\uparrow$} & 
\textbf{F1$\uparrow$} & \textbf{Per-cos$\uparrow$} & \textbf{Rea-cos$\uparrow$} \\
\midrule
\multirow{2}{*}{\shortstack[l]{SFT \\ (baseline)}} 
& \multicolumn{2}{c|}{\multirow{2}{*}{$\mathcal{L}_{sup}$}}
& Qwen2.5-VL-3B & 0.692 & 0.834 & 0.767 & 0.811 & 0.810 & 0.803 \\
& & 
& Qwen3-VL-4B & 0.816 & 0.773 & 0.775 & 0.863 & 0.855 & 0.849\\
\midrule
\multirow{8}{*}{GCL} 
& \multirow{6}{*}{$\mathcal{L}_{GCO}$} 
& \multirow{2}{*}{$ \mathcal{L}_{sup} + \mathcal{L}_{GSL}$} 
& Qwen2.5-VL-3B & 0.818 & 0.880 & 0.846 & 0.877 & 0.872 & 0.864\\
& & 
& Qwen3-VL-4B & 0.821 & 0.806 & 0.838 & 0.905 & 0.879 & 0.874\\
\cmidrule(lr){3-10}
& & \multirow{2}{*}{$ \mathcal{L}_{sup} + \mathcal{L}_{DRL}$} 
& Qwen2.5-VL-3B & 0.832 & 0.903 & 0.851 & 0.851 & 0.840 & 0.837\\
& & 
& Qwen3-VL-4B & 0.722 & 0.779 & 0.761 & 0.887 & 0.869 & 0.879\\
\cmidrule(lr){3-10}
& &\multirow{2}{*}{$ \mathcal{L}_{sup} + \mathcal{L}_{GSL} + \mathcal{L}_{DRL}$}
& Qwen2.5-VL-3B & \underline{0.908} & \underline{0.947} & \underline{0.917} & \underline{0.937} & \underline{0.936} & \underline{0.922}\\
& & 
& Qwen3-VL-4B & \underline{0.857} & \underline{0.847} & \underline{0.851} & \underline{0.943} & \underline{0.949} & \underline{0.950}\\
\cmidrule(lr){2-10}
& \multicolumn{2}{c|}{\multirow{2}{*}{$\mathcal{L}_{GCO}$ w/ AGO}} 
& Qwen2.5-VL-3B & \textbf{0.968} & \textbf{0.970} & \textbf{0.972} & \textbf{0.975} & \textbf{0.963} & \textbf{0.978}\\
& & 
& Qwen3-VL-4B & \textbf{0.914} & \textbf{0.926} & \textbf{0.927} & \textbf{0.961} & \textbf{0.969} & \textbf{0.980}\\
\bottomrule
\end{tabular}}
\end{table}

\subsection{Ablation Studies}
\label{sec:sub43}

\noindent\textbf{Performance Comparison of GCL Components}
Table~\ref{tab:ab_main} summarizes the performance of our GCL strategy. Combined with the group competitive loss, Action-F1 scores for Qwen2.5-VL-3B/Qwen3-VL-4B reach 0.908 (+31\%)/0.857 (+5\%), vastly improving upon their respective baselines (0.692 / 0.816).
AGO further refines competitive dynamics. For the 4B guide, AGO functions as a regularizer, elevating Action-F1 to 0.914 (+12\%) with marked improvements in Perception-cos and Reasoning-cos, signifying enhanced semantic robustness. Notably, the AGO enhanced 3B learner surpasses the 4B baseline across all metrics, attaining an Action-F1 of 0.968 (+40\%). This demonstrates that within GCL, a smaller model can effectively internalize complex structural knowledge from a larger counterpart, achieving generalization capabilities superior to larger models trained independently.
\begin{figure}[t]
  \centering
  \begin{subfigure}{0.49\linewidth}
    \centering
    \includegraphics[width=\linewidth]{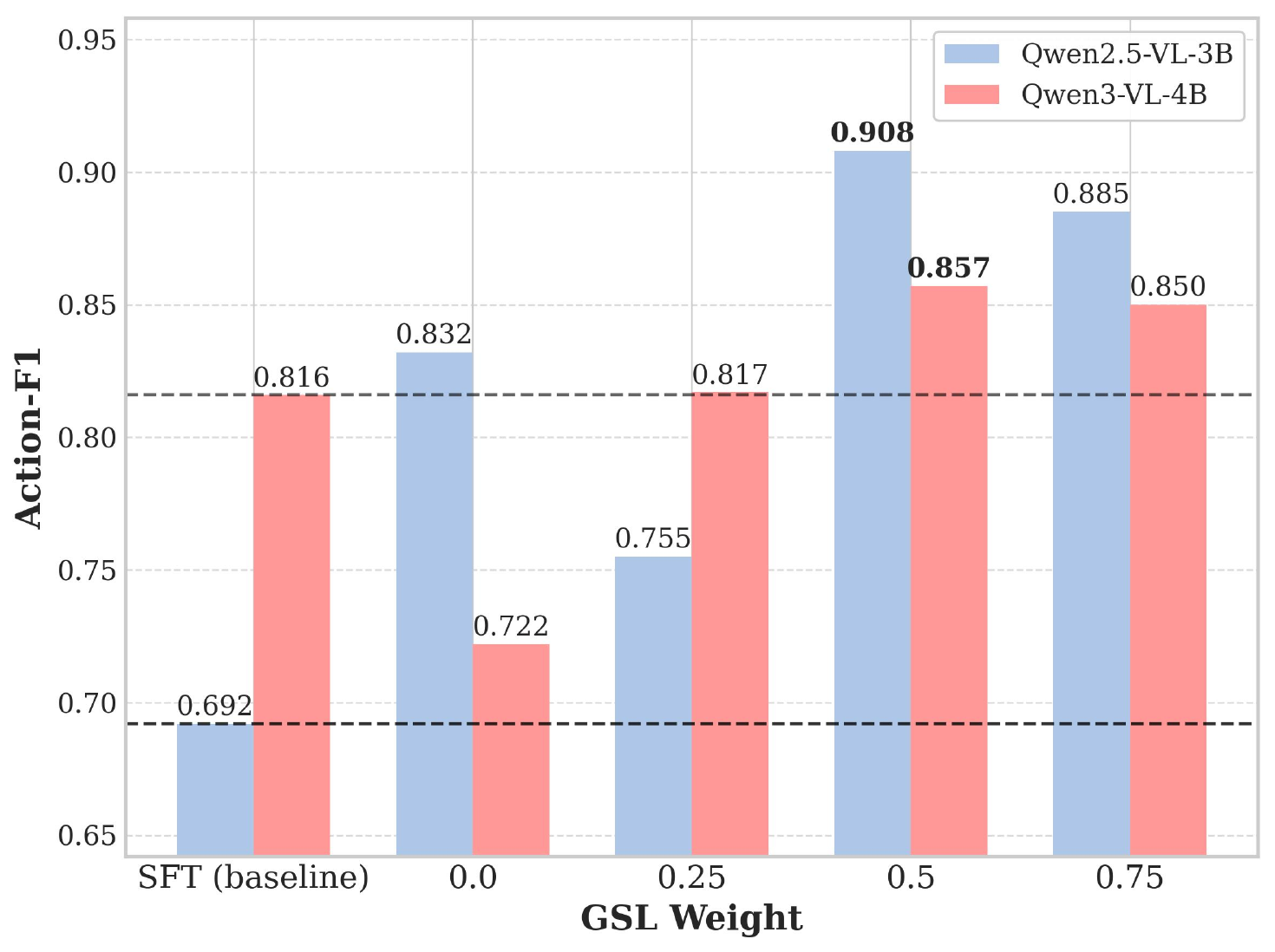}
    \caption{\textbf{Ablation study on the weight of GSL.} The guide model experiences severe degradation when explicit supervision is completely removed (weight 0.0), whereas both models achieve optimal performance at a balanced weight of 0.5.}
    \label{fig:weight_GSL}
  \end{subfigure}
  \hfill
  \begin{subfigure}{0.49\linewidth}
    \centering
    \includegraphics[width=\linewidth]{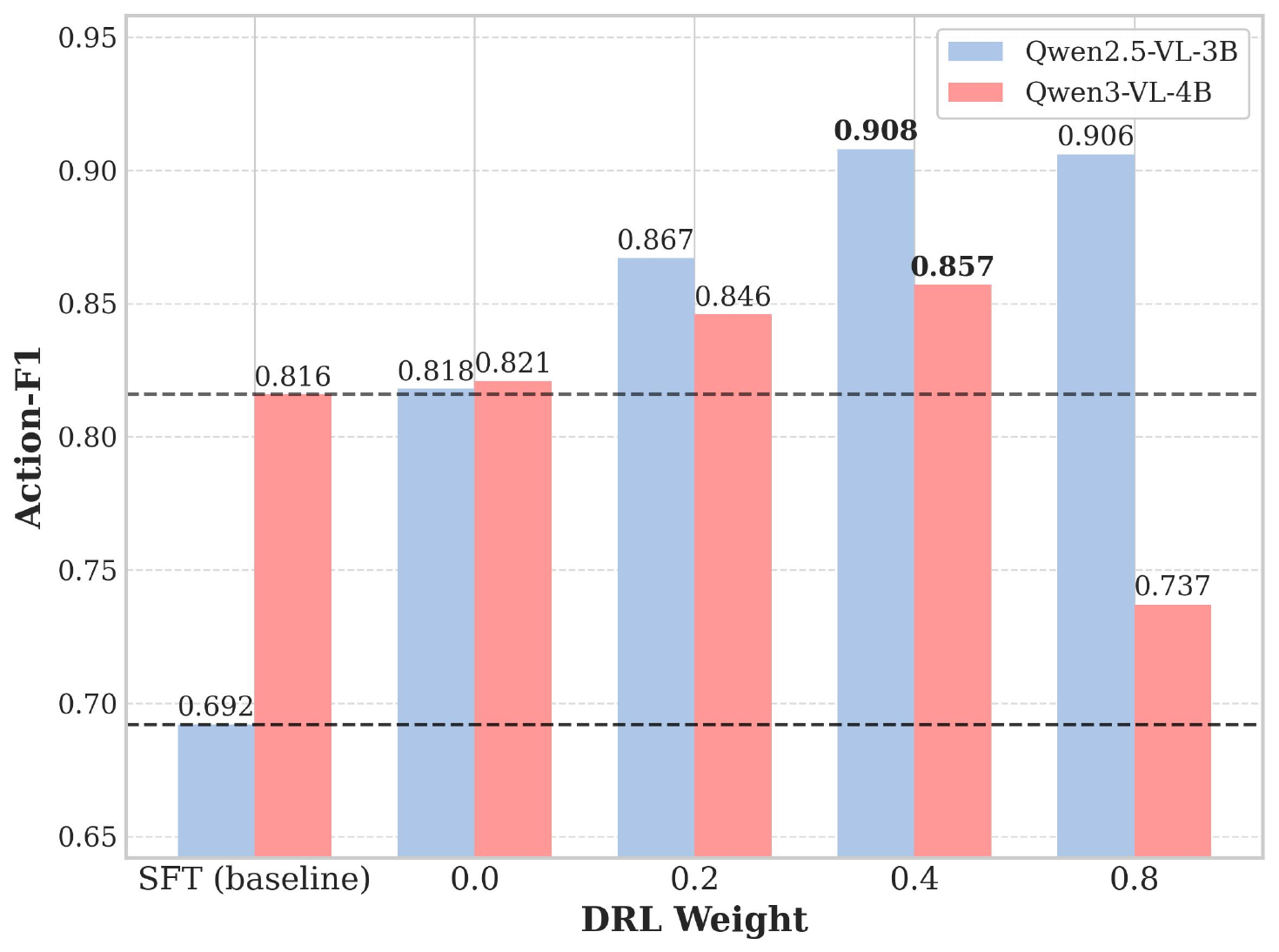}
    \caption{\textbf{Ablation study on the weight of DRL.} An excessive regularization weight (0.8) triggers a catastrophic collapse of the high capacity guide model, confirming the necessity of bounded rival alignment to maintain semantic rigidity.}
    \label{fig:weight_DRL}
  \end{subfigure}
  \caption{\textbf{Ablation Study of Loss Components in GCO.} Red and blue bars represent the learner (Qwen2.5-VL-3B) and guide (Qwen3-VL-4B) models, respectively, with bold values indicating peak performance. The left illustrates the ablation of the GSL weight, where both models achieve optimal results at 0.5. The right plot presents the ablation of the DRL weight, demonstrating that a value of 0.4 yields the highest performance for both models.}
\end{figure}

\noindent\textbf{Weight Analysis of Loss Components in GCO}
To validate the effectiveness of the core objectives (GSL and DRL), we conducted ablation studies under the Standard Setting ($\eta = 1\times 10^{-5}$).

GSL as the absolute semantic anchor in GCL, aligning optimization trajectories with semantic features. Ablation studies (Fig.~\ref{fig:weight_GSL}) demonstrate that removing GSL (weight 0.0) severely degrades the guide model (Action-F1: 0.722, SFT baseline 0.816), revealing loss of normative guidance without explicit semantic constraints. Peak performance emerges at weight 0.5 (Learner: 0.908, Guide: 0.857), confirming GSL establishes a robust optimization boundary that preserves task semantics while enabling cross-model learning. Over weighting (0.75) induces marginal decline due to compressed inter class relations and weakened distributional knowledge transfer.

DRL governs mutual alignment intensity for rival knowledge exchange (Fig.~\ref{fig:weight_DRL}). Deactivating DRL (weight 0.0) eliminates competitive dynamics, yielding only marginal guide improvement (0.821), which underscores the necessity of mutual error correction to escape local optima. Optimal alignment occurs at weight 0.4. Critically, excessive weighting (0.8) collapses the guide to 0.737 despite strong learner performance (0.906), validating that excessive alignment overwhelms the guide’s gradient stiffness, forcing mimicry of the learner’s smoothed distribution and eroding its semantic anchor role. These results establish that alignment intensity must be strictly calibrated to model capacity to prevent stronger models from being entrapped in weaker counterparts’ optimization landscapes.

\subsection{Analysis of AGO}
\label{sec:sub44}
\begin{figure*}[t!]
  \centering
  \begin{subfigure}{0.49\linewidth}
    \centering
    \includegraphics[width=\linewidth]{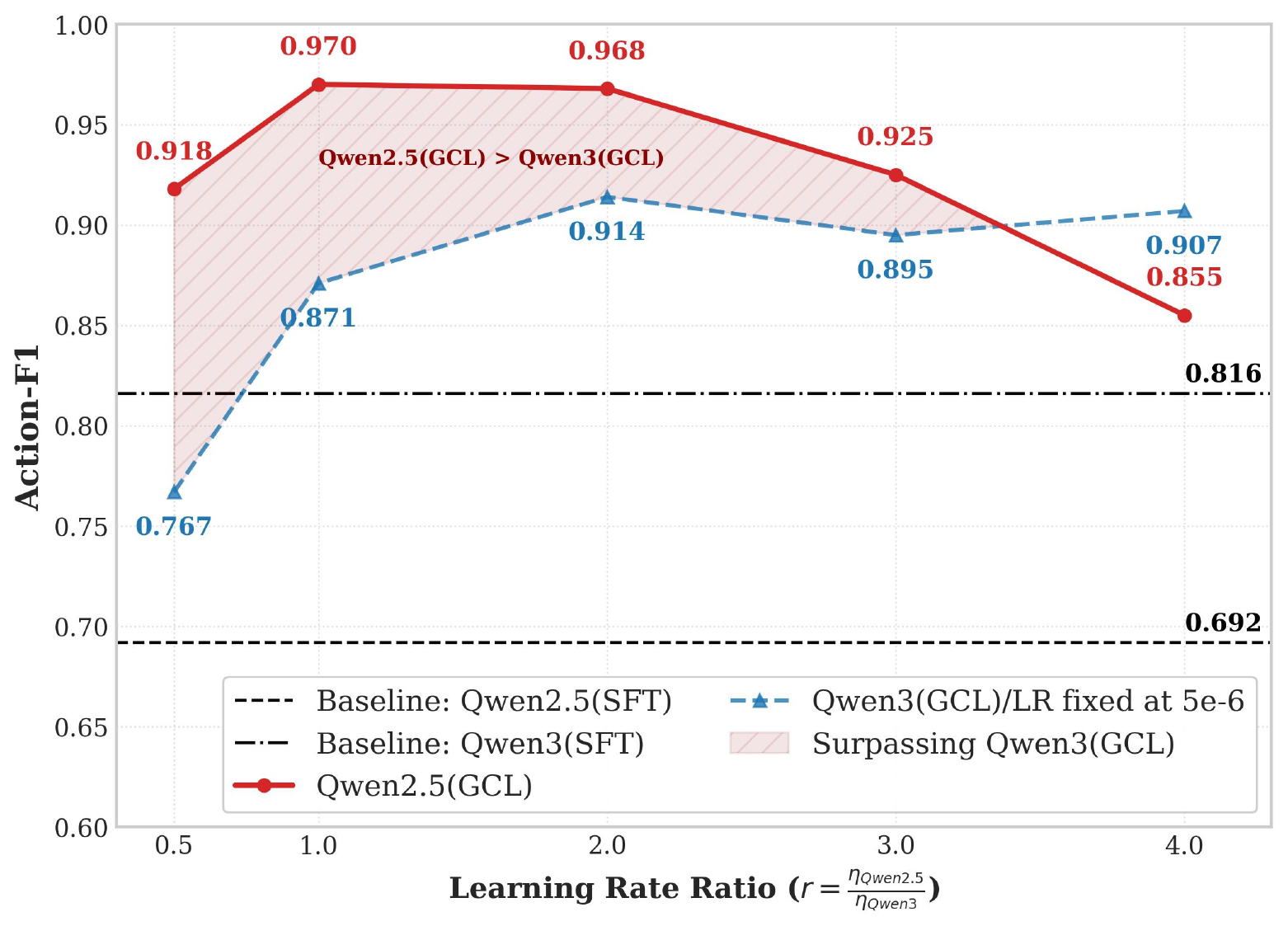}
    \caption{\textbf{High Tolerance in Large Gap Group.} In the Qwen2.5-VL-3B/Qwen3-VL-4B (SFT gap 0.124), the learner sustains peak group performance across a broad ratio range ($1.0 \le r \le 3.0$) with optimum at $r=2.0$. This wide stability window reflects the learner’s capacity to effectively bridge substantial capability disparities through calibrated update dynamics.}
    \label{fig:ago_ratio_large}
  \end{subfigure}
  \hfill
  \begin{subfigure}{0.49\linewidth}
    \centering
    \includegraphics[width=\linewidth]{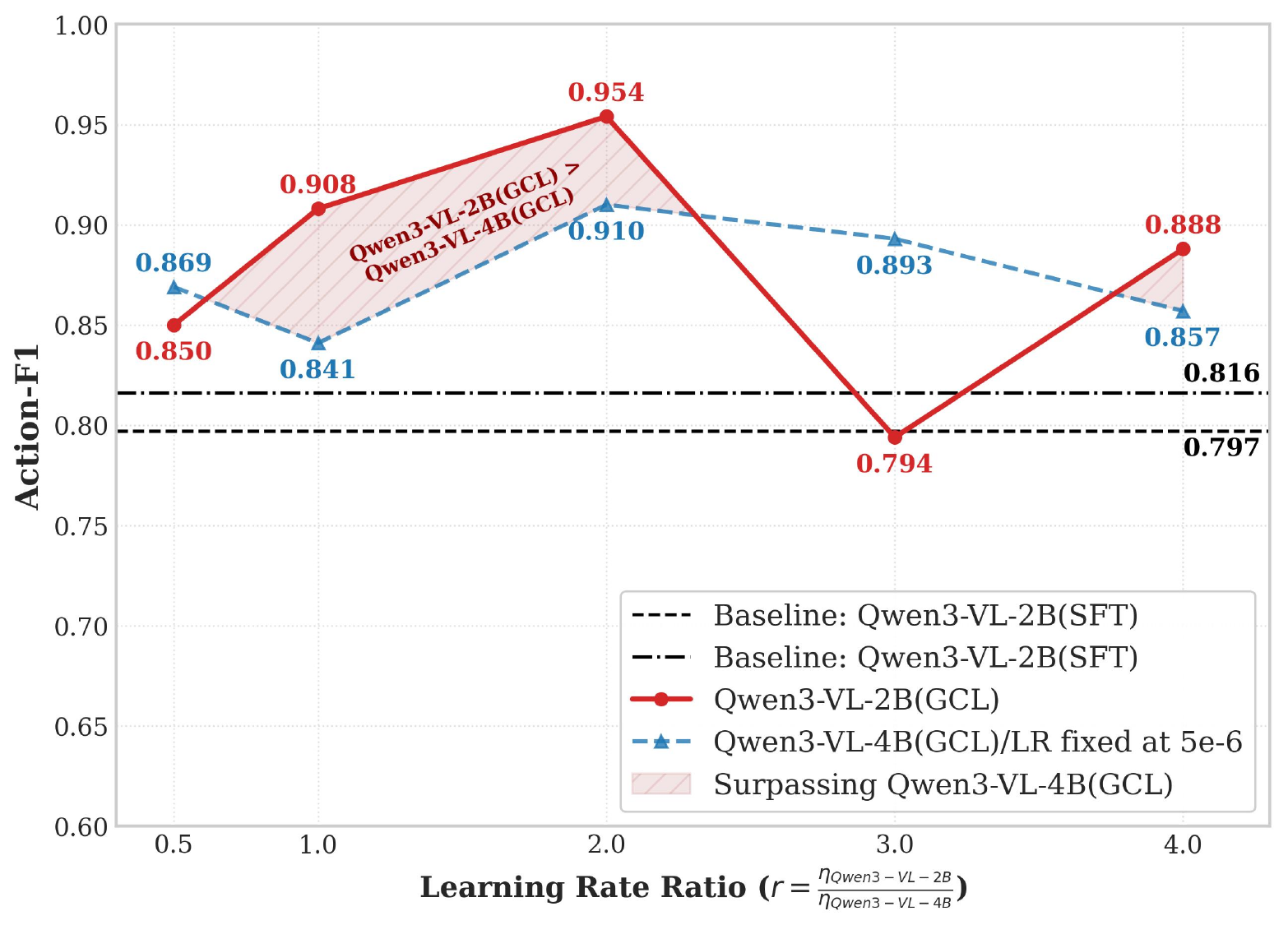}
    \caption{\textbf{Narrow Margin in Small Gap Group.} In Qwen3-VL-2B/4B, (small gap 0.019), high initial feature alignment restricts the optimal GCL to a narrow ratio window ($1.0 \le r \le 2.0$), peaking at $r=2.0$.  For overly aggressive stride (e.g., $r \ge 3.0$) causes the learner to trigger premature gradient oscillation and a performance collapse drop to 0.794.}
    \label{fig:ago_ratio_small}
  \end{subfigure}  
  \caption{\textbf{Impact of Capability Gaps on Asymmetric Learning Rate Dynamics.} Action-F1 is evaluated against the learning rate ratio ($r = \eta_{learner} / \eta_{guide}$), with $\eta_{guide}$ fixed at $\eta = 5 \times 10^{-6}$. Shaded regions indicate the Optimization Best Spot. Fig.~\ref{fig:ago_ratio_large} shows that large capability gap group tolerate aggressive learner updates ($1.0 \le r \le 3.0 $ ), enabling effective feature space alignment. In contrast, Fig.~\ref{fig:ago_ratio_small} reveals that minimal initial gaps confine the optimal window to ($1.0 \le r \le 2.0$). These confirms that GCL necessitates performance based asymmetric learning rate scheduling.}
  \label{fig:AGO}
\end{figure*}

While ablation studies validate effectiveness under standard configurations, we hypothesize that uniform learning rates and DRL temperatures constrain performance potential. To probe GCL performance boundaries and identify upper limits without exhaustive hyperparameter search, we analyze an optimized configuration. Robustness is evaluated across two model groups with distinct intrinsic capability gaps: (1) \textbf{Large gap}: Qwen2.5-VL-3B/Qwen3-VL-4B (SFT gap 0.124); (2) \textbf{Small gap}: Qwen3-VL-2B/Qwen3-VL-4B (SFT gap 0.019).

\begin{figure}[t]
\centering
\includegraphics[width=\linewidth]{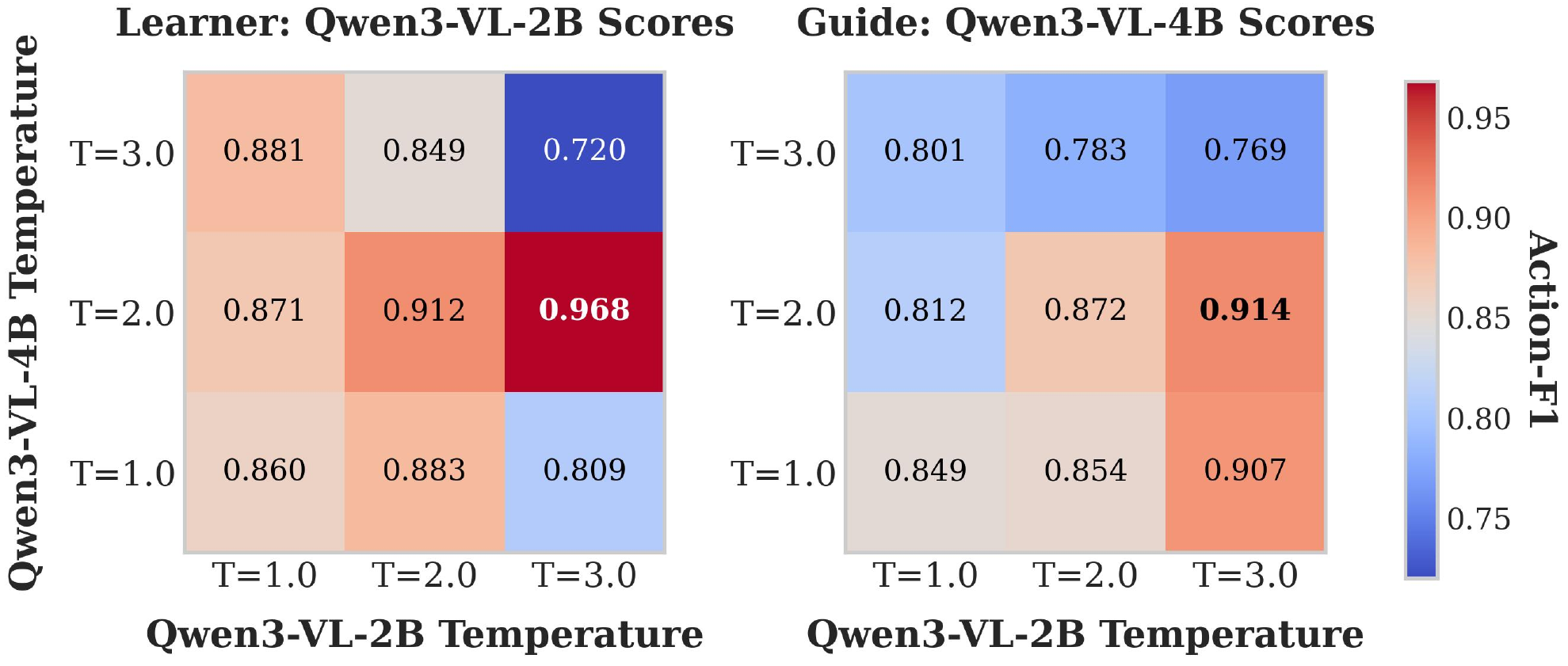}
\caption{\textbf{Optimal Asymmetric Temperature Gradients in VLMs Group.} Heatmaps of Action-F1 scores for Qwen2.5-VL-3B learner (left) and Qwen3-VL-4B guide (right) across temperature grids. Symmetric settings $\tau_{learner}=\tau_{guide}$) result in suboptimal consensus or collapse (e.g., $0.720$ at $\tau=3.0$). The global peak (learner: 0.968) occurs at $(\tau_{learner}=3.0, \tau_{guide}=2.0)$, highlighting the necessity of asymmetric thermal gradients to enhance learner plasticity through guide discriminative signaling. (\textbf{For small gap group, details provided in the Supplementary Material})}
\label{fig:heatmap}
\end{figure}

\begin{figure}[t]
    \centering
    \includegraphics[width=\linewidth]{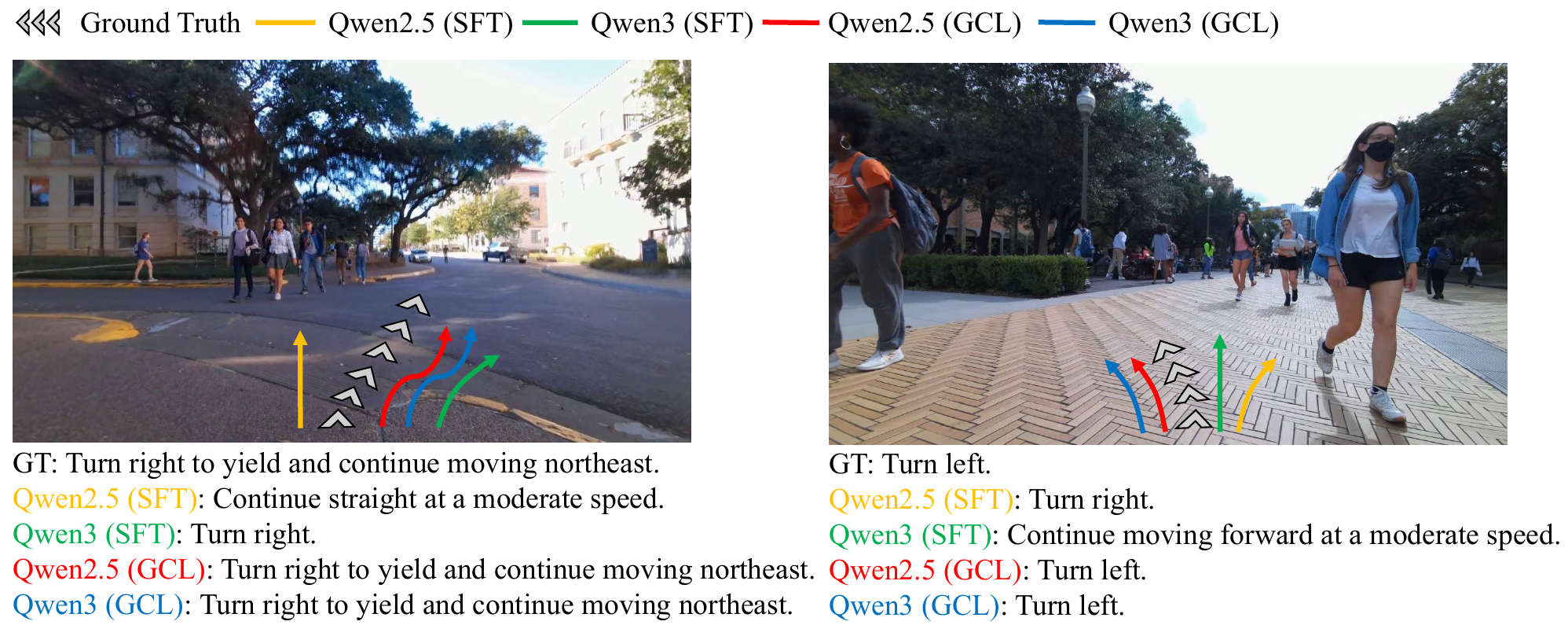}
    \caption{\textbf{Visualization Comparison.} SFT models (yellow/green) miss critical constraints and produce incorrect actions, while GCL models (red/blue) generate socially compliant instructions aligned with GT.}
    \label{fig:visual}
\vspace{0.02em}
    \centering
    \includegraphics[width=\linewidth]{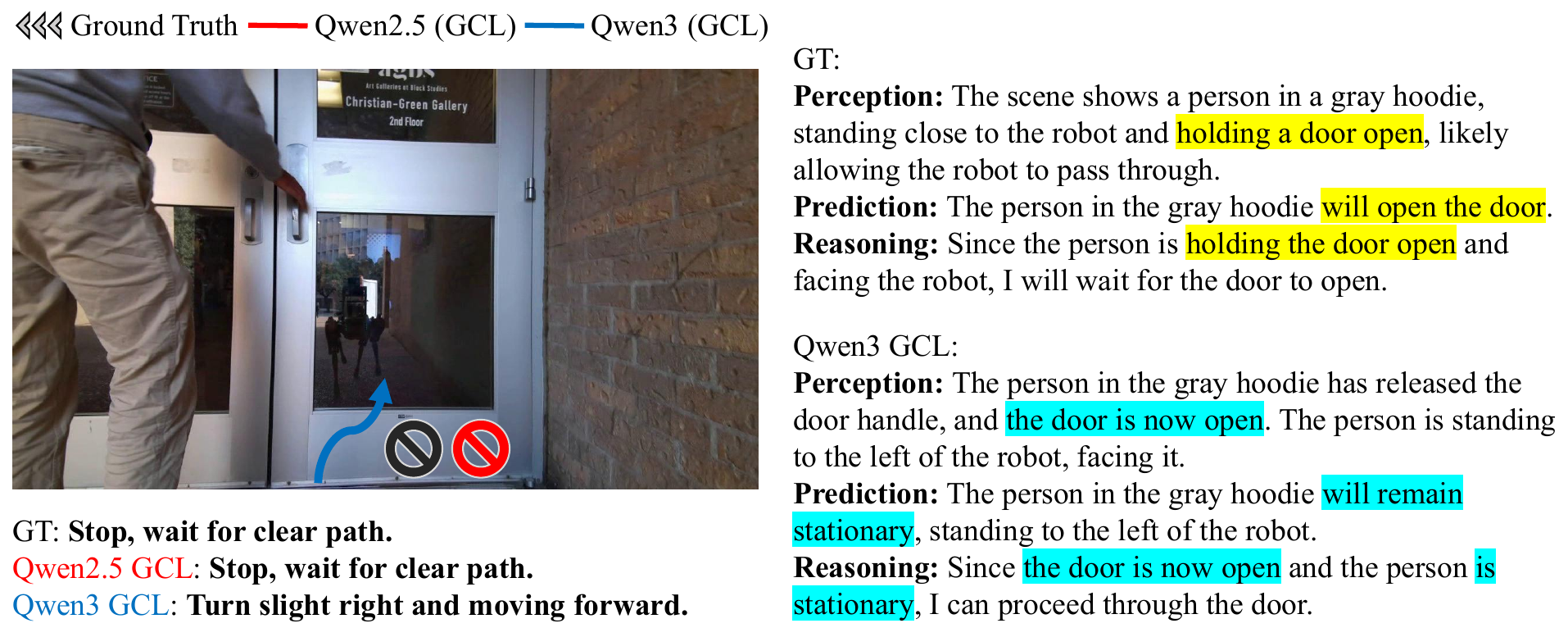}
    \caption{\textbf{Failure Case Analysis.} In this interactive navigation scenario, GT and the Qwen2.5-VL-3B learner correctly predict to stop and wait for a person holding a door, while the Qwen3-VL-4B guide misinterprets the scene and predicts safe passage.}
    \label{fig:fail}
\end{figure}

\noindent\textbf{Asymmetric Learning Rate: Stabilizing the Guide and Accelerating the Learner.} 
Balancing update rates across heterogeneous-capacity VLMs is challenging for GCL. Uniform learning rates lead to suboptimal convergence; we address this using an asymmetric scheduler that fixes the guide learning rate $\eta_{guide}$ (Qwen3-VL-4B) at $5\times10^{-6}$ as a stable anchor, while scaling the learner by $r=\eta_{learner}/\eta_{guide}$. Fig.~\ref{fig:AGO} reveals a clear ``Capacity Inversion'' phenomenon: despite lower capacity and weaker SFT baselines, learners (Qwen2.5-VL-3B/Qwen3-VL-2B) consistently surpass the guide across diverse $r$ values. This indicates that constrained guide updates prevent catastrophic forgetting and provide stable soft-label supervision, enabling the learner to explore the feature space and absorb ``dark knowledge'' beyond its architectural limits.

Further analysis shows that the optimal ratio depends on the initial capability gap. In the large gap group (Qwen2.5-VL-3B/Qwen3-VL-4B; Fig.~\ref{fig:ago_ratio_large}), the learner maintains Capacity Inversion across $1.0 \le r \le 3.0$, peaking at $r=1.0$ (Action-F1: \textbf{0.970}) and collapsing only at $r=4.0$. In contrast, the small gap group (Qwen3-VL-2B/4B; Fig.~\ref{fig:ago_ratio_small}) peaks at $r=2.0$ (Learner: \textbf{0.954}, Guide: \textbf{0.910}) but becomes unstable, with learner performance dropping to 0.794 at $r=3.0$.

These results suggest that smaller VLMs require an update stride equal to or twice that of larger models to transition from catching up to surpassing the guide within fixed epochs, supporting group-wise learning rate scheduling. When the capacity gap is small, however, large steps ($r\ge3.0$) induce \textit{Gradient Oscillation}: the learner overshoots the guide’s optimal manifold, disrupting DRL and GSL alignment. Therefore, asymmetric learning rates must be calibrated to the capacity disparity within each VLM group.

\noindent\textbf{Asymmetric Temperature: Clear Guidance and Smooth Exploration.}
Online knowledge distillation with symmetric temperatures often ignores distributional disparities across heterogeneous architectures, leading to suboptimal optimization. As shown in Fig.~\ref{fig:heatmap}, symmetric settings consistently underperform, while the best results occur with asymmetric temperatures ($\tau_{learner}=3.0$, $\tau_{guide}=2.0$). This asymmetric temperature design enhances knowledge transfer by increasing learner entropy for structural adaptation while preserving the guide’s discriminative sharpness. Experiments on both large gap (Qwen2.5-VL-3B/4B) and small gap (Qwen3-VL-2B/4B) groups (\textbf{see Supplementary Material}) show the same trend: symmetric high temperatures cause severe degradation, whereas asymmetric scheduling improves Action-F1 and stabilizes optimization. These results indicate that capacity differences require asymmetric temperature design, validating AGO as a key mechanism for robust GCL across heterogeneous lightweight VLMs.


\subsection{Visualization.} 
Fig.~\ref{fig:visual} contrasts SFT baselines and GCL trajectories against GT (gray chevrons). SFT predictions (Qwen2.5-VL-3B: yellow; Qwen3-VL-4B: green) exhibit unsafe spatial misalignments: the left scenario shows erroneous straight progression instead of a yielding right turn; the right crowded pedestrian scenario reveals unsafe maneuvers intruding into pedestrian paths. In contrast, GCL trajectories (Qwen2.5-VL-3B: red; Qwen3-VL-4B: blue) precisely align with GT, generating safe, socially compliant instructions. This confirms GCL’s significant enhancement of VLMs’ scene comprehension and critical decision-making in complex urban environments.

\subsection{Failure Analysis and Future Work.}
Despite its strong performance, GCL still faces challenges in fine-grained spatiotemporal reasoning in complex dynamic scenes. As shown in Fig.~\ref{fig:fail}, the guide model misinterprets a door interaction and predicts safe passage, revealing limitations in handling reflective surfaces and subtle human–object interactions. Future work will extend GCL to more VLM families and investigate smaller models to improve robustness and scalability.

\section{Conclusions}
In this work, we presented Group Competitive Learning (GCL), a strategy that enhances reasoning and efficiency of VLMs for robot navigation under constrained resource settings. GCL aligns both the high-level semantics and fine-grained distributions of heterogeneous model groups through a GCO, enhancing their reasoning capabilities. Through AGO, GCL addresses the dynamic mismatch problem and characterizes the performance boundaries of GCL without performing exhaustive hyperparameter search.
Experiments on two social navigation datasets demonstrate that GCL can be seamlessly integrated into the training strategies of various VLMs. Smaller learners (Qwen2.5-VL-3B and Qwen3-VL-2B) are able to surpass the larger Qwen3-VL-4B guide, improving over their SFT baselines by 40\% and 20\%, respectively. All model groups show consistent improvements regardless of capacity. Through GCL experiments using the same model (e.g., Qwen3-VL-2B), homogeneous models also achieve gains over the SFT baseline (27\%). Notably, although the 3B model initially lagged behind the 8B model with F1 scores of 0.692 and 0.755, the enhanced 3B model significantly outperforms the larger 8B baseline (by 28\%). These results indicate that our method provides meaningful benefits for social navigation and facilitates the deployment of socially aware and robust lightweight VLMs in real-world robotic systems.

%
%
\bibliographystyle{splncs04}
\bibliography{main}

\section{Additional Method Details}

In this section, we provide a derivation of the gradients for the proposed Distributional Regularization Loss ($\mathcal{L}_{DRL}$) under asymmetric temperature settings. We specifically analyze how the temperatures $\tau_A$ and $\tau_B$ interact and modulate the optimization through their gradients.

Let $z_A$ and $z_B$ be the logits produced by two VLMs. The softened probability distributions $P_{A,i}$ and $P_{B,i}$ are obtained via the softmax function with temperatures $\tau_A$ and $\tau_B$, respectively:
\begin{equation}
    P_{A,i} = \frac{\exp(z_{A,i} / \tau_A)}{\sum_j \exp(z_{A,j} / \tau_A)}, \quad P_{B,i} = \frac{\exp(z_{B,i} / \tau_B)}{\sum_j \exp(z_{B,j} / \tau_B)}
\end{equation}

The Distributional Regularization Loss (DRL) is defined as a weighted Jensen-Shannon divergence (JS), where each Kullback-Leibler ($\mathcal{D}_{KL}$) term is scaled by the square of its respective temperature to ensure gradient stability:
\begin{equation}
    \mathcal{L}_{DRL} = \frac{\tau_A^2}{2}  \mathcal{D}_{KL}(P_A || \mathcal{M}) + \frac{\tau_B^2}{2}  \mathcal{D}_{KL}(P_B || \mathcal{M})
    \label{eq:drl}
\end{equation}
where $\mathcal{M} = \frac{1}{2}(P_A + P_B)$ represents the mixed distribution.
The partial derivative with respect to $z_{A, i}$ is:
\begin{equation}
    \frac{\partial \mathcal{L}_{DRL}}{\partial z_{A, i}} = \frac{\tau_A^2}{2} \frac{\partial \mathcal{D}_{KL}(P_A || \mathcal{M})}{\partial z_{A, i}} + \frac{\tau_B^2}{2} \frac{\partial \mathcal{D}_{KL}(P_B || \mathcal{M})}{\partial z_{A, i}}
\end{equation}
Using the softmax derivative $\frac{\partial P_{A,j}}{\partial z_{A,i}} = \frac{1}{\tau_A} P_{A,j}(\delta_{ij} - P_{A,i})$, we analyze the two components: 
\begin{itemize}
    \item Self-term: Expanding the derivative of $\mathcal{D}_{KL}(P_A || \mathcal{M})$ yields:
    \begin{equation}
    \begin{split}
        \frac{\tau_A^2}{2} \frac{\partial \mathcal{D}_{KL}(P_A || \mathcal{M})}{\partial z_{A, i}} 
        &= \frac{\tau_A}{2} P_{A, i} \left( \log \frac{P_{A, i}}{\mathcal{M}_i} - \mathcal{D}_{KL}(P_A || \mathcal{M}) \right. \\
        &\quad \left. - \frac{P_{A, i}}{2\mathcal{M}_i} + \mathbb{E}_{P_A} \left[ \frac{P_A}{2\mathcal{M}} \right] \right)
    \end{split}
    \end{equation}
    where $\mathbb{E}_{P_A}$ denotes expectation of probability distribution A.
    \item Cross-term: Since $P_B$ is fixed relative to $z_A$, the gradient flows only through the mixed distribution $\mathcal{M}$ in the denominator of the KL term:
    \begin{equation}  
    \frac{\tau_B^2}{2} \frac{\partial \mathcal{D}_{KL}(P_B || \mathcal{M})}{\partial z_{A, i}} = - \frac{\tau_B^2}{2\tau_A} P_{A, i} \left( \frac{P_{B, i}}{2\mathcal{M}_i} - \mathbb{E}_{P_A} \left[ \frac{P_B}{2\mathcal{M}} \right] \right)
    \end{equation}
\end{itemize}

To simplify the expression, let $V_i = \frac{P_{B, i}}{P_{A, i} + P_{B, i}} = \frac{P_{B, i}}{2\mathcal{M}_i}$ represent the relative contribution of Cross-term to the mixture at index $i$. Consequently, $\frac{P_{A, i}}{2\mathcal{M}_i} = 1 - V_i$. Substituting $V_i$ into the Self-term:
\begin{equation}
    \text{Self-term} = \frac{\tau_A}{2} P_{A, i} \left( \log \frac{P_{A, i}}{\mathcal{M}_i} - \mathcal{D}_{KL} \right) - \frac{\tau_A}{2} P_{A, i} \left( (1 - V_i) - \mathbb{E}_{P_A}[1 - V] \right)
\end{equation}
Using the linearity of expectation $\mathbb{E}_{P_A}[1 - V] = 1 - \mathbb{E}_{P_A}[V]$, the constant terms cancel out:
\begin{equation}
    \text{Self-term} = \frac{\tau_A}{2} P_{A, i} \left( \log \frac{P_{A, i}}{\mathcal{M}_i} - \mathcal{D}_{KL} \right) + \frac{\tau_A}{2} P_{A, i} \left( V_i - \mathbb{E}_{P_A}[V] \right)
\end{equation}
Combining the adjusted Self-term and the Cross-term:
\begin{equation}
    \frac{\partial \mathcal{L}}{\partial z_{A, i}} = \frac{\tau_A}{2} P_{A, i} \left( \log \frac{P_{A, i}}{\mathcal{M}_i} - \mathcal{D}_{KL} \right) + \left( \frac{\tau_A}{2} - \frac{\tau_B^2}{2\tau_A} \right) P_{A, i} (V_i - \mathbb{E}_{P_A}[V])
\end{equation}
The coefficient in the second term can be rewritten as:
\begin{equation}
    \frac{\tau_A}{2} - \frac{\tau_B^2}{2\tau_A} = \frac{\tau_A^2 - \tau_B^2}{2\tau_A} = - \frac{\tau_B^2 - \tau_A^2}{2\tau_A}
\end{equation}
This results in the final simplified form for the gradient with respect to $z_{A, i}$:
\begin{equation}
\begin{split}
    \frac{\partial \mathcal{L}}{\partial z_{A, i}} 
    &= \underbrace{\frac{\tau_A}{2} P_{A, i} \left( \log \frac{P_{A, i}}{\mathcal{M}_i} - \mathcal{D}_{KL}(P_A || \mathcal{M}) \right)}_{\textbf{Basic Alignment}} \\
    &\quad - \underbrace{\frac{\tau_B^2 - \tau_A^2}{2\tau_A} P_{A, i} \left( \frac{P_{B, i}}{P_{A, i} + P_{B, i}} - \mathbb{E}_{P_A} \left[ \frac{P_B}{P_A + P_B} \right] \right)}_{\textbf{Asymmetric Shift Force}}
\end{split}
\end{equation}

Comparing the results for $z_{A,i}$ and $z_{B,i}$, we observe an Asymmetric Shift Force modulated by the temperature difference $\Delta \tau^2 = |\tau_A^2 - \tau_B^2|$. 
When $(\tau_B^2 - \tau_A^2) = 0$, the DRL formulation (Eq.~\ref{eq:drl}) degenerates to the standard JS divergence.
When $(\tau_B^2 - \tau_A^2) < 0$, the DRL formulation (Eq.~\ref{eq:drl}) changes to the general distillation loss.
When $(\tau_B^2 - \tau_A^2) > 0$, smaller models $B$, with their limited parameter capacity, are highly susceptible to overfitting and often produce non-smooth output probability distributions. Assigning a high temperature $\tau_{high}$ to the smaller model explicitly smooths its originally sharp output probabilities. This not only serves as a powerful regularization mechanism but also compels the smaller model to learn soft inter-class relationships. In contrast, larger models $A$ possess abundant parameters and strong representational capacity, yet remain prone to overconfidence on training data. Setting a low temperature $\tau_{low}$ for the larger model enables accurate prediction while introducing mild regularization through the guiding influence of the smaller model’s high temperature outputs, thereby effectively mitigating overconfidence.

\section{Additional Experiment Details}

This section we provide the experimental implementation details and presents supplementary experimental results.

\subsection{Additional Implementation Details}
To ensure a fair comparison, all experiments are conducted on 8 NVIDIA RTX 8000 GPUs, utilizing DeepSpeed ZeRO-3 for distributed training with the Adam optimizer. The learning rate follows a cosine annealing schedule, with all models trained for 10 epochs and a warmup ratio of 0.1.

\subsection{Additional Main Results Details}
\begin{table*}[t]
\centering
\caption{\textbf{Performance of GCL across 12 competitive group on MUSON.} $\eta_{guide}$ and $\eta_{learner}$ are allocated to SFT baseline superiority. The performance hierarchy from lowest to highest is: \textbf{Qwen3-VL-2B (0.805) $<$ Qwen2.5-VL-3B (0.811) $<$ Qwen2.5-VL-7B (0.829) $<$ Qwen3-VL-4B (0.863) $<$ Qwen3-VL-8B (0.873).} Meanwhile, temperature parameters ($\tau_L$, $\tau_G$) are assigned solely based on parameter size (larger capacity = 2.0, smaller capacity = 3.0). $F1$ denotes Action-F1. Bold text indicates optimal performance of the model.}
\label{tab:muson_main}
\resizebox{\textwidth}{!}{
\begin{tabular}{l|c|c|c}
\toprule
\multicolumn{4}{l}{
\makecell[l]{\textbf{Regime A: Reshaping Mode} \\
\textbf{Capacity\textsubscript{L} < Capacity\textsubscript{G},\quad Small $\rightarrow$ Large,\quad $\bm{\eta_{learner} = 1 \times10^{-5}}$ $\bm{\eta_{guide} = 5 \times10^{-6}}$ }}} \\
\midrule
\textbf{Model Group (Learner / Guide)} & $\mathbf{\tau_L}$ / $\mathbf{\tau_G}$ & \textbf{Baseline ($F1_L$ / $F1_G$)} & \textbf{GCL ($F1_L$ / $F1_G$)} \\
\midrule
Qwen3-VL-2B / Qwen2.5-VL-3B  & \textbf{3.0 / 2.0} & 0.805 / 0.811 & \textbf{0.989} {\color{teal}(+0.184)} / 0.974 {\color{teal}(+0.163)} \\
Qwen3-VL-2B / Qwen3-VL-4B    & \textbf{3.0 / 2.0} & 0.805 / 0.863 & 0.963 {\color{teal}(+0.158)} / 0.969 {\color{teal}(+0.106)} \\
Qwen2.5-VL-3B / Qwen3-VL-4B  & \textbf{3.0 / 2.0} & 0.811 / 0.863 & 0.975 {\color{teal}(+0.164)} / 0.961 {\color{teal}(+0.098)} \\
Qwen3-VL-2B / Qwen2.5-VL-7B  & \textbf{3.0 / 2.0} & 0.805 / 0.829 & 0.907 {\color{teal}(+0.102)} / 0.943 {\color{teal}(+0.114)} \\
Qwen2.5-VL-3B / Qwen2.5-VL-7B& \textbf{3.0 / 2.0} & 0.811 / 0.829 & 0.966 {\color{teal}(+0.155)} / 0.930 {\color{teal}(+0.101)} \\
Qwen3-VL-2B / Qwen3-VL-8B    & \textbf{3.0 / 2.0} & 0.805 / 0.873 & 0.969 {\color{teal}(+0.164)} / 0.979 {\color{teal}(+0.106)}  \\
Qwen2.5-VL-3B / Qwen2.5-VL-8B& \textbf{3.0 / 2.0} & 0.811 / 0.873 & \textbf{0.983} {\color{teal}(+0.172)} / 0.977 {\color{teal}(+0.104)} \\
Qwen3-VL-4B / Qwen3-VL-8B    & \textbf{3.0 / 2.0} & 0.863 / 0.873 & 0.960 {\color{teal}(+0.097)} / 0.968 {\color{teal}(+0.095)}\\
Qwen2.5-VL-7B / Qwen3-VL-8B  & \textbf{3.0 / 2.0} & 0.829 / 0.873 & 0.945 {\color{teal}(+0.116)} / \textbf{0.981} {\color{teal}(+0.108)} \\
Qwen3-VL-2B / Qwen3-VL-2B    & \textbf{3.0 / 2.0} & 0.805 / 0.805 & 0.875 {\color{teal}(+0.070)} / 0.893 {\color{teal}(+0.088)} \\
Qwen2.5-VL-3B / Qwen2.5-VL-3B& \textbf{3.0 / 2.0} & 0.811 / 0.811 & 0.911 {\color{teal}(+0.100)} / 0.891 {\color{teal}(+0.080)} \\
\midrule
\multicolumn{4}{l}{
\makecell[l]{\textbf{Regime B: Extraction Mode} \\
\textbf{Capacity\textsubscript{L} > Capacity\textsubscript{G},\quad Large $\rightarrow$ Small,\quad $\bm{\eta_{learner} = 1 \times10^{-5}}$ $\bm{\eta_{guide} = 5 \times10^{-6}}$ }}} \\
\midrule
\textbf{Model Group (Learner / Guide)} & $\mathbf{\tau_L}$ / $\mathbf{\tau_G}$ & \textbf{Baseline ($F1_L$ / $F1_G$)} & \textbf{GCL ($F1_L$ / $F1_G$)} \\
\midrule
Qwen2.5-VL-7B / Qwen3-VL-4B  & \textbf{2.0 / 3.0} & 0.829 / 0.863 & \textbf{0.952} {\color{teal}(+0.123)} / \textbf{0.987} {\color{teal}(+0.124)} \\
\bottomrule
\end{tabular}
}
\end{table*}
We adopt identical experimental settings to those employed on the SNEI dataset and evaluate Group Competitive Learning (GCL) performance on the MUSON dataset across 12 competitive groups for fair comparison (As shown in Table~\ref{tab:muson_main}).

Experimental results demonstrate that, through the GCL strategy, the smaller learners (Qwen2.5-VL-3B and Qwen3-VL-2B) outperform the larger guides (Qwen3-VL-4B and Qwen3-VL-8B). Furthermore, uniform performance gains are observed across all 12 competitive groups, irrespective of model capacity.
\begin{figure}[t]
\centering
\includegraphics[width=\linewidth]{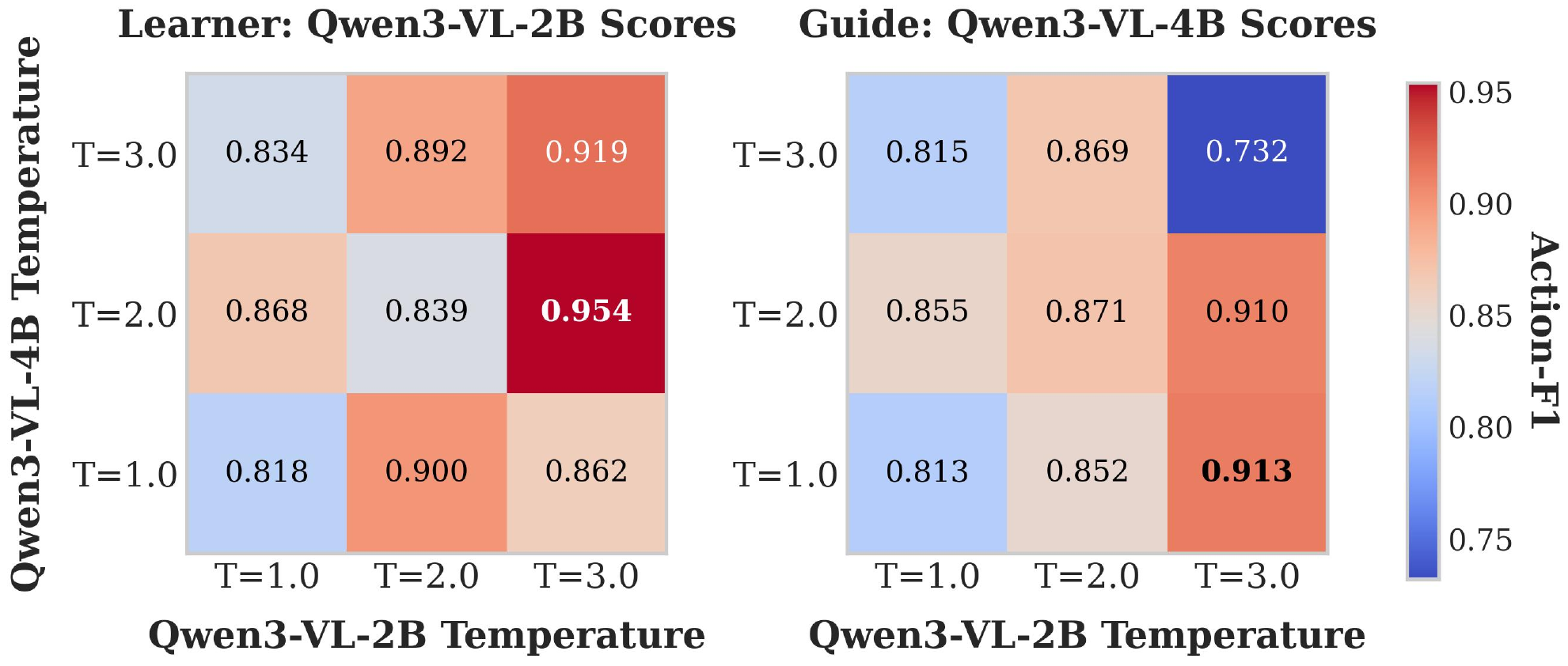}
\caption{\textbf{Asymmetric Temperature in Small Gap Group.} Heatmaps of Action-F1 scores for Qwen3-VL-2B learner (left) and Qwen3-VL-4B guide (right) across temperature grids. The global peak (learner: 0.954) occurs at $(\tau_{learner}=3.0, \tau_{guide}=2.0)$. Symmetric configuration ($\tau_{learner}=\tau_{guide}$) leads to suboptimal performance and may even induce severe performance degradation (guide: 0.732).}
\label{fig:heatmap2}
\end{figure}

\subsection{Additional Asymmetric Temperature Details}
We present the experimental results of asymmetric temperature configurations for the competitive group with small gap. As illustrated in the Fig.~\ref{fig:heatmap2}, the symmetric temperature setting yields suboptimal performance and notable degradation (guide: 0.732). The globally optimal configuration is identified as $(\tau_{learner}=3.0, \tau_{guide}=2.0)$. This substantiates that, irrespective of model capacity disparity, asymmetric temperature ($(\tau_{learner}=3.0, \tau_{guide}=2.0)$) scheduling consistently enhances the performance of competitive groups within the GCL.

\end{document}